# Automatic Processing of Proper Names in Texts

## Francis Wolinski[1] [2] Frantz Vichot[1] Bruno Dillet[1]

1 Informatique CDC            2 LAFORIA
Caisse des Dépôts et Consignations   Université de Paris VI
France                        France
E-mail: {wolinski,vichot,dillet}@icdc.fr

## Abstract

This paper shows first the problems raised by proper names in natural language processing. Second, it introduces the knowledge representation structure we use based on conceptual graphs. Then it explains the techniques which are used to process known and unknown proper names. At last, it gives the performance of the system and the further works we intend to deal with.

## 1    Introduction

The Exoseme system [6, 7] is an operational application which continuously analyses the economic flow from Agence France Presse (AFP). AFP, which covers the current economic life of the major industrialised countries, transmits on average 400 dispatches per day on this flow. Their content is drafted in French in a journalistic style. Using this flow, Exoseme feeds various users concerning precise and varied subjects, for example, rating announcements, company results, acquisitions, sectors of activity, observation of competition, partners or clients, etc. 50 such themes have currently been developed. They rely on precise filtering of dispatches with highlighting of sentences for fast reading.

Exoseme is composed of several modules : a morphological analyser, a proper name module, a syntactical analyser, a semantic analyser and a filtering module. The proper name module has two goals : segmenting and categorising proper names. During the whole processing of a dispatch, the proper name module is involved in three different steps. First, it segments proper names during the morphological analysis. Second, it categorises proper names during the semantic analysis. Third, it is invoked by the filtering module to supply some more information needed for routing the dispatch.

The proper name module is based on different techniques which are used to detect and categorise proper names depending on whether they are known or unknown. Some of these techniques are taken out of existing systems but they have been unified and completed in constructing this single operational module. Besides some innovative techniques for desambiguating known proper names using the context have been implemented.

## 2    Problems raised by proper names in NLP

In the AFP flow, proper names constitute a significant part of the text. They account for approximately one third of noun groups and half the words used in proper names do not belong to the French vocabulary (e.g. family names, names of locations, foreign nouns). In addition, the number of words used in constructing proper names is potentially infinite.

The first step of the processing is segmentation, i.e. accurate cutting-up of proper names in the text; the second step is categorisation, i.e. the attribution to each proper name of a conceptual category (individual, company, location, etc.). It should be noted that segmentation and categorisation are processed differently depending on whether the proper name is known or unknown.

### 2.1    Segmentation of proper names

The segmentation of proper names enables the syntactical analyser to be relieved, particularly in the case of long proper names which contain grammatical markers (e.g. prepositions, conjunctions, commas, full stops). As illustrated in [4], segmentation firstly prevents long proper names from undertaking pointless analyses. For example, for Caisse de Crédit Agricole du Morbihan the analyser will provide two interpretations depending on whether Morbihan is attached to Crédit Agricole or to Caisse. Moreover, proper names often constitute agrammatical segments that sometimes confuse the syntactical analyser. For exam-ple, in the sentence `The director of Dollfus, Mieg and Cie has announced positive results`, the analyser has difficulties in finding that `The director is the subject of` `announce` if it does not know the company `Dollfus, Mieg and Cie`. In the Exosome process, the Sylex syntactical analyser [3] delegates the segmentation of these agrammatical gaps to our proper name module.

Segmentation of known proper names has already been studied and is treated in some systems such as NameFinder [5]; segmentation of unknown proper names based on pattern matching is implemented in several systems [1, 2, 4, 9]; the morphological matching of acronyms is described in [11].

## 2.2 Categorisation of proper names

Once the segmentation has been achieved, categorisation of proper names is necessary for the semantic analyser. Categorisation maps proper names into a set of concepts (e.g. human being, company, location). The very nature of proper names contributes widely to the understanding of texts. The semantic analyser must be able to use the various categories of proper names as semantic constraints which are complementary for the understanding of texts. For example, in the filtering theme of acquisitions, the sentence `Express group intends to sell Le Point for 700 MF` indicates a sale of interests in the newspaper `Le Point`. Whereas the following sentence, which is grammatically identical to the preceding one, `Compagnie des Signaux intends to sell TVM430 for 700 MF` indicates only a price for an industrial product.

Categorisation of unknown proper names has already been studied as well. Particularly, categorisation of unknown proper names is automatically acquired in pattern matching techniques quoted in previous section; rules using the context of proper names in order to categorise them are also implemented in [2, 9].

In our system, these ontological categories are extended to attributes needed by the semantic analyser or the filtering module. For instance, proper names may have different attributes such as city, rating agencies, sector of activity, market, financial indexes, etc.

## 3 Representation of proper names

We will see that the proper name module requires a large amount of information concerning proper names, their forms, their categories, their attributes, the words of which they are composed, etc. This information must be able to be enriched in order to include additional processes, and accessible in order to be shared by several processes. We use a representation system similar to conceptual graphs [10], the flexibility of which effectively gives expressiveness, reusability and the possibility of further development. It enables indispensable and heterogeneous data to be memorised and used in order to process proper names.

For a given proper name, its category and its various attributes are directly represented in the form of a conceptual graph. For example, our knowledge base contains the graphs of Figure 1. This simple representation will be completed in the subsequent sections. We are going to show how each encountered problem uses the information of the knowledge base and may add its own information to it.

The final result is a large knowledge base including 8,000 proper names sharing 10,000 forms, based on 11,000 words. There are also 90 attributes of proper names or words. Each new filtering theme may be a special case and its implementation may lead to introduce additional attributes into the knowledge base. The adopted representational formalism enables these additions to be made without leading to substantial modifications of its structure.

## 4 Processing known proper names

Firstly, we recognise the proper names in which we are directly interested in order to allocate to them attributes which are required for subsequent processes. We also seek to recognise the most frequent proper names (e.g. country, cities, regions, statesmen) in order to segment them and categorise them correctly.

### 4.1 Immediate recognition

The first idea which comes to mind is to memorise the proper names as they are encountered in the dispatches and to allocate to them the attributes. All this information is stored in the knowledge base which contains, for example :

- ``New'' + ``York'' → PN → location

- ``Société'' + ``Générale'' → PN → bank

- ``Standard'' + ``and'' + ``Poor's'' → PN → rating agency

The knowledge base is thus structured on the model showed in Figure 2. And subsequently, recognition of the proper name in the text occurs through simple pattern matching.

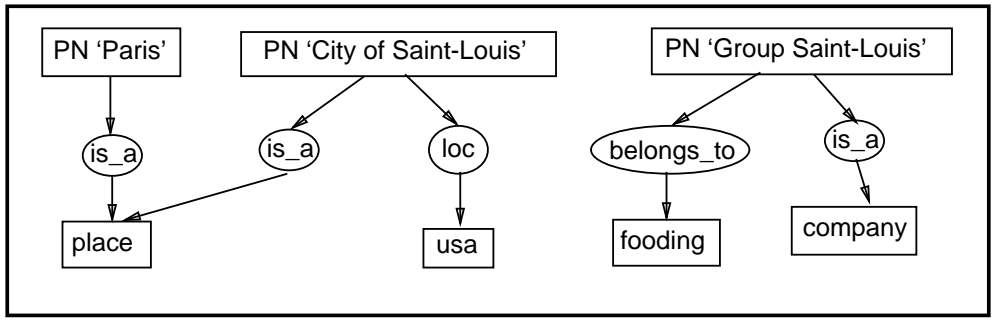

Figure 1: Representation of Proper Names

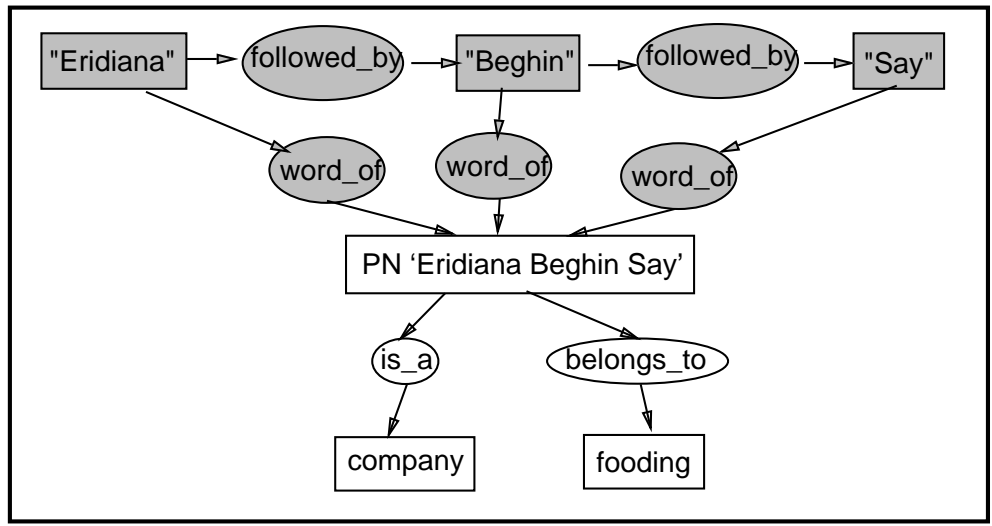

Figure 2: Words composing Proper Names

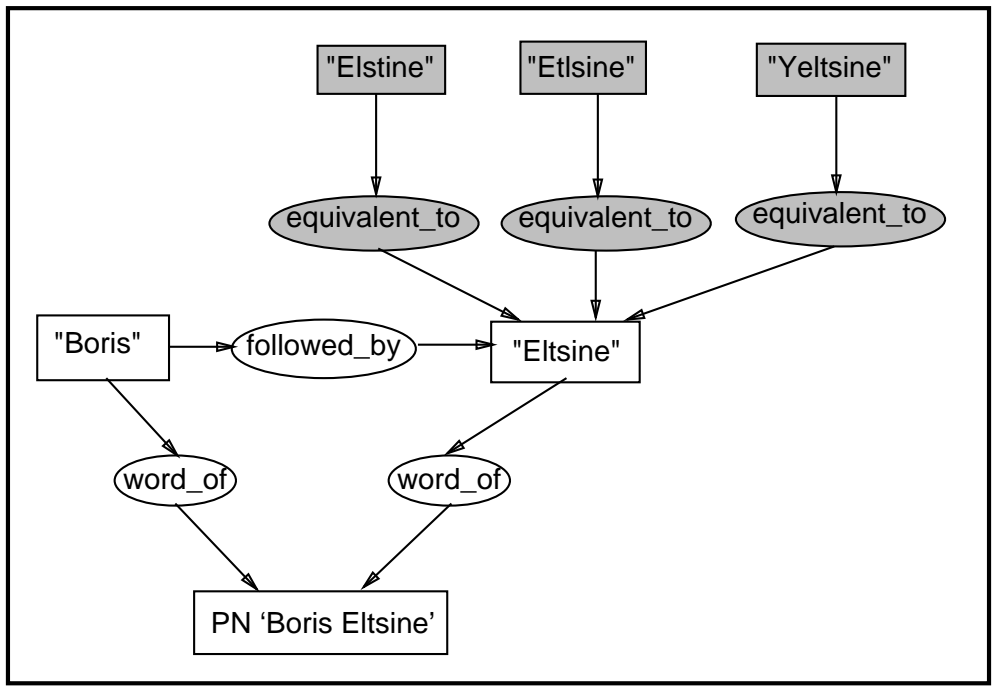

Figure 3: Equivalent Words

## 4.2 "Equivalent" words

However, words making up proper names accept many slippages which result from abbreviations, translation, common faults, etc. For example :

- `Standard and Poor's` :
  `Standard and Poors, Standard et Poor's`

- `Société Générale` :
  `Soc. gen., Sté générale`

- `Boris Eltsine` :
  `Boris Elstine, Boris Etlsine, Boris Yeltsine`

In order to avoid listing pointlessly all the forms that a proper name can take, through slippages of its words, certain variations in the recorded form are authorised. To this end, slippages in a given word are grouped around an "equivalent". This technique, which has been developed in the NameFinder system [5], under the term "alternative" words, enables to make a correspondence with different forms likely to appear.

Equivalent words are expressed in the knowledge base through a relationship. For example, our base contains the graph of Figure 3.

## 4.3 Synonymous proper names

However, one can use very different proper names to designate a given reality. For example, we can find simple synonyms such as `Hexagone` for `France` or `Rue d'Antin` for `Paribas`. This notion is similar to alternative names in [5]. Dispatches also contain more or less complex transformations, that it can be difficult to derive from the standard form, such as `NewYork` and `NY` for `New York`, or indeed `SetP` and `S-Poors` for `Standard and Poor's`.

Once again, in order to avoid listing pointlessly the attributes for all the necessary proper names, the forms of synonymous proper names are grouped around a single reference to which the various attributes are allocated. This grouping enables the various references memorised to be represented, and their attributes to be factorised. The knowledge base is modified according to the enriched model showed in Figure 4.

## 4.4 Disambiguating proper names

When a user is interested in a given proper name, it is not sufficient to look for it through the dispatches since a simple selection on this name frequently produces homonyms. Such interference, which is annoying for users, reflects the limitations of traditional keyword systems. In the AFP flow, for example, the form `Saint-Louis` may designate equally well:

- the capital of Missouri,

- a french group in the food production industry,

- les Cristalleries de Saint Louis,

- a small town in Bas-Rhin province,

- an hospital in Paris,

The crucial problem posed is to succeed in disambiguating this type of forms. Or, in other words, in determining, or at least in delimiting, the denoted reference.

### 4.4.1 Disambiguating through the local context

Exploration of the local context using the proper name can in certain cases enable a choice to be made between these various references. If the text speaks of `St-Louis (Missouri)`, only the first interpretation will be adopted, if the knowledge base contains the information that `Saint-Louis` is in the United States, and if a rule is able to interpret the affixing of a parenthesis. We are currently working on this delicate aspect in order to unify all the rules we have accumulated for resolving concrete cases. We are aware that these types of inference are comparable to the micro-theories of the Cyc project [8] in which the need for a great amount of information is the main thesis.

We will see in section 5.2.1 that the local context may categorise an unknown proper name and therefore it may help to desambiguate an ambiguous known proper name. For instance, if the text speaks of `the mayor of St-Louis`, the company and hospital can certainly be ruled out.

### 4.4.2 Disambiguating through the global context

Abbreviations of proper names are another, much more frequent, source of ambiguities. Depending on the context, `la Générale` may designate `Société Générale`, `Compagnie Générale des Eaux` or indeed `Générale de Sucrière`. Similarly, acronyms, which are almost always common to several proper names, constitute an extreme form of abbreviation. We thus discover from time to time new organisations which share the acronym `CDC` with `Caisse des Dépôts et Consignation`.

In general, ambiguous forms are not used on their own in dispatches, and other non-ambiguous forms appear. Their presence consequently enables the ambiguity to be removed. If the proper names `Saint Louis` and `Hôpital Saint Louis` appear in a single dispatch, for example, the reference corresponding to the hospital will have more forms than each of the others and will thus be the only one adopted.

Consequently, when there is an interest in an individual reference and the corpus has revealed homonyms, we record them in the knowledge base. We link them with the individual reference in order to be able to manage the ambiguities.

Nevertheless, when the ambiguity is unable to be removed, we choose the most frequent interpretation, but the user is told of the doubtful nature of our choice. In the dispatch title "`Saint Louis: results up`", for example, the proper name `Saint Louis` is processed as the food production group, which is the most frequent case, although it could equally well designate `les Cristalleries`.

# 5 Processing unknown proper names

The preceding techniques tackled the problem of the variability of known proper names. However, although many proper names appear frequently, others appear only once. Even if the constituted knowledge base is very comprehensive, it is absolutely impossible to record all potential proper names. We have therefore to deal with unknown proper names.

## 5.1 Prototypes of proper names

As fully explained in [2], some proper names are constructed according to prototypes which enable them to be categorised through their appearance alone. For example :

- `known-first-name + unknown-upcase-word` $\longrightarrow$ human being (e.g. André Blavier)

- `unknown-upcase-word + company-legal-form` $\longrightarrow$ company (e.g. Kyocera Corp)

- `unknown-upcase-word + ''-sur-'' + unknown-upcase-word` $\longrightarrow$ location (e.g. Condé-sur-Huisne)

Furthermore, certain categories of proper names accept traditional extensions which it is also possible to detect. For example :

- `known-human-being + human-title` $\longrightarrow$ human being (e.g. Kennedy Jr)

- `known-company + company-activity` $\longrightarrow$ company (e.g. Honda Motor)

- `known-company + ''-'' + known-location` $\longrightarrow$ company (e.g. IBM-France)

- `known-human-being + company-activity` $\longrightarrow$ company (e.g. Bernard Tapie Finance)

Lastly, such extensions may be combined, e.g. "`Siam Nissan Automobile Co Ltd`" is probably a subsidiary of `Nissan`.

These prototypes enable both to segment and categorise proper names. Of course, they do not constitute infallible rules (for example, `Guy Laroche` is a company while its prototype makes one believe it is a human being) but they give correct results in a large majority of cases.

In order to use these prototypes, we build a rulebase for detecting and extending proper names. Moreover, we add some attributes to the existing words in our knowledge base (e.g. first names, legal company forms, company activities). For example, it contains the graph of Figure 5.

## 5.2 Other techniques of categorisation

Nevertheless, a prototype is not always enough to categorise a proper name. In particular, an isolated proper name does not enable one to infer its category directly. For example, who can say simply on sight of the proper name that `Peskine` is an individual, `Fibaly` a company and `Gisenyi` a town ?

### 5.2.1 Categorisation through the local context

However, the text often contains elements enabling one to deduce the category of a proper name [2]. To this end, rules using the local context give good results. For example :

- apposition of an individual's position : `Peskine, director of the group`,

- name complement typical of a company : `the shareholders of Fibaly`

- name complement typical of a location : `the mayor of Gisenyi`.

These rules once again require that certain words from the knowledge base are marked by individual attributes. For example, the word "`mayor`" has both the following attributes :

- `human-being-apposition` : (e.g. Chirac, mayor of the town)

- `location-name-complement` : (e.g. the mayor of Royan)

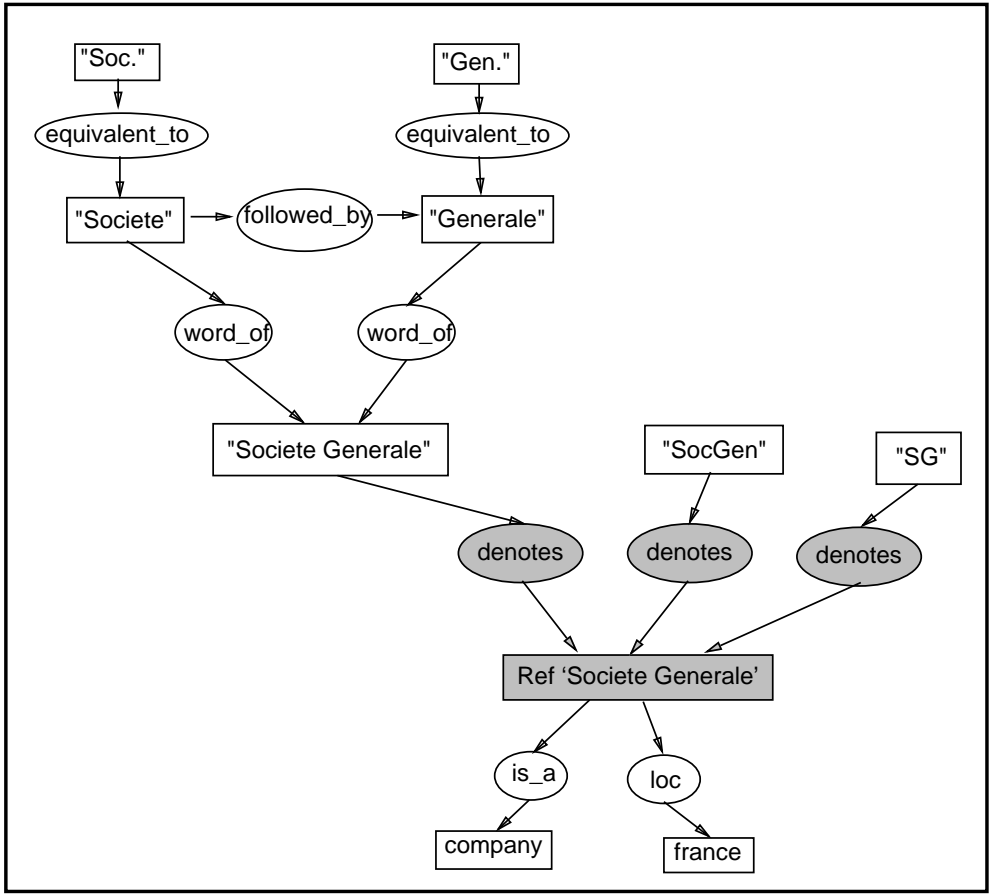

Figure 4: Synonymous Proper Names

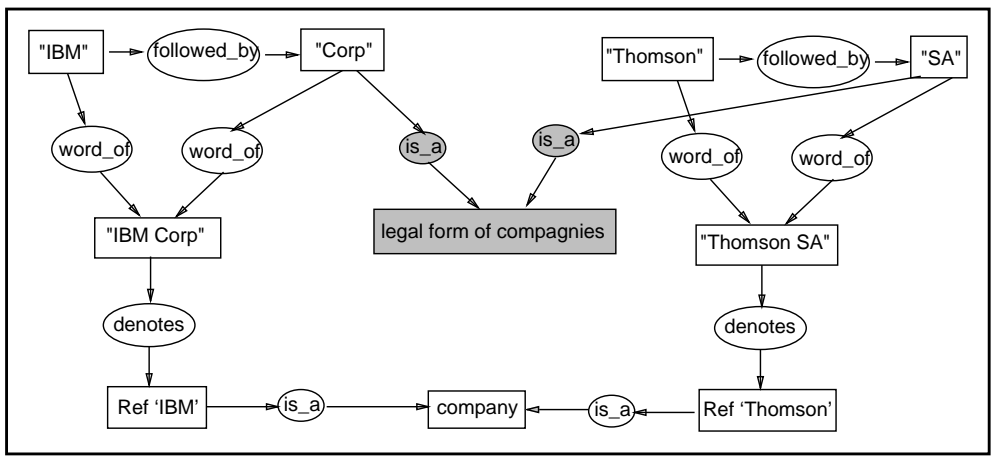

Figure 5: Words and Proper Names Attributes

### 5.2.2 Categorisation through the global context

However, the local context of a proper name does not necessarily enable one to infer its category. For instance, the mere radical of a proper name (e.g. family name, main company) is often used later in the text instead of the full name. The company `Kyocera Corp`, for example, may be designated by the single word `Kyocera` in the remainder of the text.

Consequently, for each unknown proper name, we look to see whether it does not appear in another proper name in the text. In this case, we establish a link between these two proper names in order to transfer the attributes of the recognised proper name to this new proper name. However, one should always beware since different proper names sometimes share the same radical : `Mr Mitterand` and `Mrs Mitterand`, or again `Mr Bolloré` and `Bolloré Group`. Although, in the most frequent cases, we resolve this well-known problem but as in [11] we do not have a general solution.

### 5.3 Matching acronyms

Acronyms occur frequently in AFP dispaches. On one hand, the linguistical construction of the corresponding text of acronyms may be relatively complex. On the other hand, in some case, the relatively simple morphological construction of acronyms may be treated with a simple pattern matching with the corresponding text. Moreover, acronyms are widespread ambiguous forms of which it is unthinkable to list all cases and we have seen in section 4.4.2 that desambiguation of proper names needed to memorize all potential homonyms. Therefore, a process for dealing with acronyms will first segment these unknown proper names and second desambiguate these potential homonyms.

In general, when an acronym is introduced in a text, its complete form is given using parentheses. For example :

- `International Primary Aluminium Institute (IPAI)`
- `AIEA (Agence Internationale de l' Energie Atomique)`
- `Centre de recherche, d'études et de documentation en économie de la santé (CREDES)`

As observed in [11], it is possible to explore the local structure of the parentheses in order to determine whether the acronym corresponds to the complete form and, if so, the acronym and the full name are propagated throughout the remainder of the text. Some words (e.g. articles, prepositions) may be jumped when matching up acronyms and text. For example, the acronym `SBF` of `Société des Bourses Françaises` omits the preposition "des", while the acronym `BDF` of `Banque de France` keeps the "de". In order for our processing module to recognise these words, we allocate a special attribute to them in the knowledge base.

This simple and effective technique enables most of the acronyms introduced to be processed correctly. Only foreign acronyms accompanied by their translation are not processed.

## 6 Results and prospects

Built for an operationnal system which filters in real time AFP dispatches, we have presented the module for the automatic processing of proper names. This module unifies and completes known techniques which enable to segment and categorise proper names. Particularly, we have explained our innovative technique for disambiguating known proper names and its relationship with the techniques for categorising unknown proper names and for matching acronyms. Our system currently detects 90% of proper names in AFP dispatches and categorises 85% of them correctly. The full Exoseme process is undertaken in approximately 14 seconds per dispatch on a SUN SPARC 10, i.e. in 1,400 words/minute approximately.

We consider continuing with our work relating to the exploration of the local context (Cf. 4.4.1 and 5.2.1) in two complementary directions. From the grammatical point of view, our exploration of the context is incomplete. For example, we do not categorise the unknown proper name in a complex case such as `Its Belgian subsidiary specialising in flat products Nokia`. From the semantic point of view, we do not use all the contextual data. For example, the sentence `The company already serves Houston, Saint-Louis and Dallas` should be sufficient to disambiguate `Saint-Louis`. We are currently accumulating examples in which the local context enables certain proper names to be categorised and/or to be disambiguated. Our next step will consist in tightening cooperation with the following layers in order to use the grammatical and semantic data they provide in the whole process.

## Aknowledgements

We would like to thank André Blavier, Jean-François Perrot and Jean-Marie Sézérat and the referees for their comments on versions of this paper.

# References

[1] ANDERSEN P., HAYES P., HUETTNER A., SCHMANDT L., NIRENBURG I., WEINSTEIN S. 1992 *Automatic Extraction from Press Releases to Generate News Stories*, ANLP '92

[2] COATES-STEPHEN S. 1992 *The Analysis and Acquisition of Proper Names for Robust Text Understanding*, Ph.D. Department of Computer Science of City University, London, England

[3] CONSTANT P. 1991 *Analyse syntaxique par couche*, Thèse Télécom Paris, France

[4] JACOBS P., RAU L. 1993 *Innovations in text interpretation*, Artificial Intelligence 63

[5] HAYES PH. 1994 *NameFinder : Software that find names in Text*, RIAO '94 New York

[6] LANDAU M.-C., SILLION F., VICHOT F. 1993 *Exoseme : A Thematic Document Filtering System*, Avignon '93

[7] LANDAU M.-C., SILLION F., VICHOT F. 1993 *Exoseme : A Document Filtering System Based on Conceptual Graphs*, ICCS '93

[8] LENAT D., GUHA R. 1990 *Building large Knowledge-based Systems : Representation and Inference in the Cyc Project*, Addison-Wesley

[9] MCDONALD D. 1994 *Trade-off Between Syntactic and Semantic Processing in the Comprehension of Real texts*, RIAO '94 New York

[10] SOWA J. 1984 *Conceptual Structures. Information Processing Mind and Machines*, Addison-Wesley

[11] WACHOLDER N., RAVIN Y., BYRD R. 1994 *Retrieving Information from Full Text Using Linguistic Knowledge*, IBM Research Report